\pdfoutput=1

\documentclass[11pt]{article}

\usepackage{acl}

\usepackage{times}
\usepackage{latexsym}

\usepackage[T1]{fontenc}

\usepackage[utf8]{inputenc}
\usepackage{microtype}

\usepackage{booktabs,arydshln}
\usepackage{fixltx2e}
\usepackage{amsmath}
\usepackage{multirow}
\usepackage{graphicx}
\usepackage{spverbatim}
\usepackage{newfloat}
\usepackage{subcaption}

\newcommand{\dn}{{\usefont{T1}{ppl}{m}{n}Head-to-Tail}}

\title{{\dn}: How Knowledgeable are Large Language Models (LLMs)?\\
A.K.A. Will LLMs Replace Knowledge Graphs?}

\author{Kai Sun, Yifan Ethan Xu, Hanwen Zha, Yue Liu, Xin Luna Dong \\
Meta Reality Labs \\
  \texttt{\{sunkaicn,ethanxu,hwzha,yuei,lunadong\}@meta.com} \\}

\newcommand{\todo}[1]{}

\newcommand{\eg}{{e.g.}}
\newcommand{\ie}{{i.e.}}
\newcommand{\resp}{{resp. }}
\newcommand{\vs}{{vs. }}

\DeclareFloatingEnvironment[
    fileext=los,
    listname={List of Prompts},
    name=Prompt,
    placement=tbhp,
]{prompt}

\begin{document}
\maketitle

\begin{abstract}
Since the recent prosperity of Large Language Models (LLMs), there have been interleaved discussions regarding how to reduce hallucinations from LLM responses, how to increase the factuality of LLMs, and whether Knowledge Graphs (KGs), which store the world knowledge in a symbolic form, will be replaced with LLMs. In this paper, we try to answer these questions from a new angle: {\em How knowledgeable are LLMs?}

To answer this question, we constructed {\dn}, a benchmark that consists of $18$K %
question-answer (QA) pairs regarding head, torso, and tail facts in terms of popularity. %
We designed an automated evaluation method and a set of metrics that closely approximate the knowledge an LLM confidently internalizes. %
Through a comprehensive evaluation of 16 publicly available LLMs, we show that existing LLMs are still far from being perfect in terms of their grasp of factual knowledge, especially for facts of {\em torso-to-tail} entities.

\end{abstract}

\section{Introduction}

Pre-trained large language models (LLMs), such as ChatGPT\footnote{\url{https://openai.com/blog/chatgpt}}, GPT-4~\cite{openai2023gpt4}, and Llama~2~\cite{touvron2023llama2}, have demonstrated impressive capabilities in internalizing knowledge and responding to common inquiries~\cite{ouyang2022training,openai2023gpt4}. Nevertheless, these models often lack knowledge of nuanced, domain-specific details and are susceptible to hallucinations~\cite{bang2023multitask}, underscoring the significant challenges of increasing the {\em factuality} of LLMs and minimizing {\em hallucinations} from LLM responses. Conversely, the rise of LLMs has sparked debates on whether Knowledge Graphs (KGs), which store real-world factual knowledge in triplet form (subject, predicate, object), will be replaced with LLMs. This paper tries to answer these questions from a new angle: {\em How knowledgeable are LLMs?}

\begin{figure}[t!]
\centering
\includegraphics[width=0.45\textwidth]{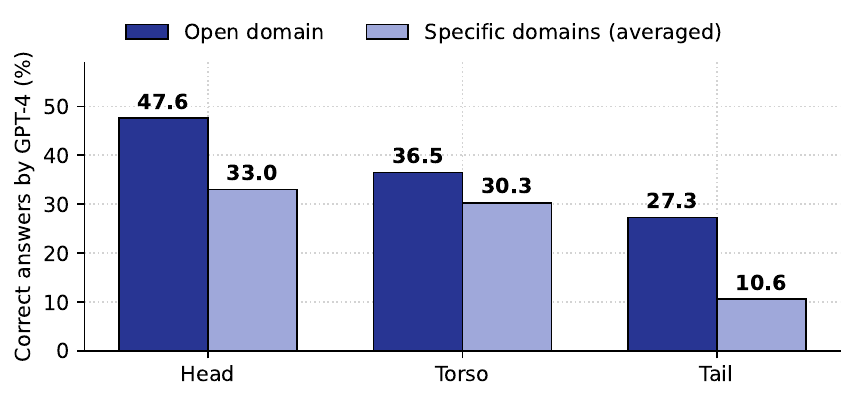}
\scriptsize

\begin{tabular}{p{0.45\textwidth}}

\\
\textbf{Example questions where GPT-4 gives incorrect answers} \\
\toprule
\textbf{Movie} \\
\noalign{\vskip 0.5ex}\hdashline\noalign{\vskip 0.5ex}
\textbf{Question:} What profession does Tj Singh (known for John Carter (2012)) have? \\
\textbf{Ground Truth:} Visual effects \\
\textbf{GPT-4:} Actor \\
\midrule
\textbf{Book} \\
\noalign{\vskip 0.5ex}\hdashline\noalign{\vskip 0.5ex}
\textbf{Question:} Who authored Choke (published in 1996)? \\
\textbf{Ground Truth:} Stuart Woods \\
\textbf{GPT-4:} Chuck Palahniuk \\
\midrule
\textbf{Academics} \\
\noalign{\vskip 0.5ex}\hdashline\noalign{\vskip 0.5ex}
\textbf{Question:} Where did Josef Kittler receive the Ph.D. (thesis: Development and application of pattern recognition techniques.)? \\
\textbf{Ground Truth:} University of Cambridge, UK \\
\textbf{GPT-4:} University of Surrey \\
\midrule
\textbf{Open} \\
\noalign{\vskip 0.5ex}\hdashline\noalign{\vskip 0.5ex}
\textbf{Question:} What college is the sister college of Trinity College, Oxford? \\
\textbf{Ground Truth:} Churchill College, Cambridge \\
\textbf{GPT-4:} Balliol College \\

\bottomrule
\end{tabular}

\caption{The question-answering accuracy of GPT-4 decreases in the order of head, torso, and tail entities on the {\dn} benchmark, and is only 31\% on average.}
\label{tab:overview}
\end{figure}

Finding answers to these questions is not easy. First, it is hard to directly ``query'' the knowledge embedded in an LLM---hallucination can be due to lack of knowledge but can also be caused by dysfunction of the generative model even if the knowledge is already parameterized in the model. We approximate the amount of knowledge in an LLM by its accuracy in answering simple-formed questions, such as {\em ``where was the basketball player Michael Jordan born?''}; in addition, we ask the LLM to generate brief answers and admit ``unsure'' when its confidence is low. We chose this proxy because we found LLMs are normally very good at understanding simple-formed questions and produce consistent answers when regenerating answers, especially if asked to be brief (Section~\ref{sec:rq4}). 

Second, there is no ready-to-use benchmark that either well represents distributions of user's interest (the query logs for major LLMs or search engines are not publicly available) or well represents the uniform distribution of the world knowledge (even the largest knowledge graphs admit sparsity of knowledge, especially towards non-popular facts). To address this challenge, we construct a benchmark of $18$K %
QA pairs that cover various domains and various relationships in these domains. We bucket entities and relationships to {\em head, torso,} and {\em tail} according to how {\em popular} they are (details in Section~\ref{sec:data}) and randomly sample from each bucket; as such, we call our benchmark {\dn}. This benchmark facilitates us to achieve a comprehensive view of how knowledgeable LLMs are regarding each bucket.

Through the {\dn} benchmark and the experimental methodology, we answer the following three research questions (RQs): 
    \begin{itemize}
        \item[\textbf{RQ1:}] How reliable are LLMs in answering factual questions? (Section~\ref{sec:rq1})
        \item[\textbf{RQ2:}] Do LLMs perform equally well on head, torso, and tail facts? (Section~\ref{sec:rq2})
        \item[\textbf{RQ3:}] Do normal methods that improve LLMs, such as model size increase and instruction tuning, help LLMs to be more knowledgeable? (Section~\ref{sec:rq3})
    \end{itemize}

As shown in Figure~\ref{tab:overview}, our analysis demonstrates a consistent decline in the performance of LLMs, following the order of head, torso, and tail entities, confirming our hypothesis that LLMs contain more head knowledge where training data abound. 
Surprisingly, even for the top-0.5\% popular entities in popular domains such as {\em Movie}, the evaluated LLMs, at best, provide accurate answers for only $\sim$60\% of the questions in the benchmark. 
Normal methods that enhance LLMs do not necessarily make them more knowledgeable, highlighting the need for more effective approaches to increase LLMs' factuality.

Our main contributions are as follows:

\begin{itemize}
    \item[\textbf{(i)}] We introduce {\dn}, the first benchmark focused on comprehensively assessing the effectiveness of LLMs in incorporating factual knowledge encompassing the head, torso, and tail portions of knowledge graphs (Section~\ref{sec:qapairgen}). {\dn} will be  available at \url{https://github.com/facebookresearch/head-to-tail}.%
    \item[\textbf{(ii)}] We present an evaluation methodology accompanied by metrics designed to assess the factuality of LLMs. Our metrics allow us to distinguish hallucination and missing answers, and our evaluation method, whereas entirely automated, proves to be reliable and robust (Section~\ref{sec:metrics}-\ref{sec:evalmethod}).
    \item[\textbf{(iii)}] We conducted a comprehensive evaluation and quantified the factuality of 16 LLMs regarding head, torso, and tail facts to answer the research questions (RQ1--RQ3) (Section~\ref{sec:experiment}). In light of these findings, we envision the future of knowledge graphs and outline a research landscape aimed at improving the overall factual reliability of LLMs (Section~\ref{sec:disc}).
\end{itemize}

\section{The {\dn} Benchmark}
\label{sec:data}

\todo{figure}
We now describe the {\dn} benchmark, the metrics, and our evaluation methodology.

\subsection{QA pair generation}
\label{sec:qapairgen}

\par{\noindent\textbf{Domains and data sources.}} 
To cover a broad range of knowledge, we used the DBpedia knowledge graph~\cite{soren2007dbpedia}, where the knowledge originates from Wikipedia~\cite{wikipedia}. We used a cleaned version of the English snapshot from December 1, 2022.\footnote{\url{https://databus.dbpedia.org/dbpedia/mappings/mappingbased-objects}}

To better understand LLM performance on particular domains, we also selected three domains where public data are easily accessible. 
\begin{itemize}
    \item \textbf{Movie:} We used a snapshot of {\em IMDb}\footnote{\url{https://developer.imdb.com/non-commercial-datasets/}} from May 21, 2023.
    \item \textbf{Book:} We used the data of {\em Goodreads} scraped in 2017 released by \citet{wan2018item}.
    \item \textbf{Academics:} We used a snapshot of {\em MAG}~\cite{sinha2015overview} from September 13, 2021 and {\em DBLP}\footnote{\url{https://dblp.org/}} from May 10, 2023.
\end{itemize}

\begin{table*}[ht!]
\centering
\tiny
\begin{tabular}{lrrrrrrrr}
\toprule
 & \multicolumn{2}{c}{\textbf{IMDb}} & \bf Goodreads & \multicolumn{3}{c}{\textbf{MAG}} & \bf DBLP & \bf DBpedia  \\
 & Title & Person & Book & Article & Conference & Journal & Scholar & - \\
\cmidrule(rl){2-3} \cmidrule(rl){4-4} \cmidrule(rl){5-7} \cmidrule(rl){8-8} \cmidrule(rl){9-9}

\bf Head & 767 (\phantom{0}0.01)& 34,903 (\phantom{0}0.48)& 3,150 (\phantom{0}2.31)& 1,827,710 (\phantom{0}0.70)& 257 (\phantom{0}1.63)& 225 (\phantom{0}0.46)& 79,521 (\phantom{0}2.44)& 103,564 (\phantom{0}1.30)\\
\bf Torso & 4,113 (\phantom{0}0.05)& 87,645 (\phantom{0}1.21)& 7,304 (\phantom{0}5.35)& 9,386,034 (\phantom{0}3.60)& 965 (\phantom{0}6.12)& 1,266 (\phantom{0}2.58)& 500,778 (15.36)& 1,255,113 (15.77)\\
\bf Tail & 7,536,482 (99.94)& 7,111,496 (98.31)& 126,134 (92.35)& 249,311,539 (95.70)& 14,550 (92.25)& 47,546 (96.96)& 2,680,704 (82.20)& 6,600,206 (82.93)\\

\bottomrule
\end{tabular}
\caption{The number (\%) of head, torso, and tail entities. The distribution follows the power law.}
\label{tab:entitypartition}
\end{table*}

\par{\noindent\textbf{Entities.}} An important contribution of the {\dn} benchmark is the bucketing of head, torso, and tail entities, decided by the {\em popularity} of the entities (we will also discuss how the popularity of the predicates affect results in Section~\ref{sec:rq2}). We use two ways to approximate popularity: {\em traffic} and {\em density}. When there is traffic information, such as views and votes, we conveniently use traffic to measure the popularity; otherwise, we use density as a proxy, such as the number of facts or authored works about the entity. We often observe a correlation between density and traffic (\eg, the more popular a person is, the more we know about her), but as we will see soon from the benchmark statistics (Table~\ref{tab:entitypartition}), they can still lead to slightly different distributions of head, torso, and tail. We give details on how we decide the popularity of different types of entities from each data source in Appendix~\ref{sec:popularitymeasure}. %

We bucketed head, torso, and tail entities in three steps. First, we sorted the entities by their popularity, measured as above. Second, for each entity, we computed the cumulative popularity score up to the top-1 entity in the sorted list. Third, we bucketed the entities such that head entities comprise entities whose cumulative popularity score is up to 1/3 of that of all entities, torso entities comprise entities with cumulative scores ranging from 1/3 to 2/3, and tail entities from 2/3 to 1. (See Appendix~\ref{sec:entitybucketingexample} for an example.) We determined the partitioning separately for different entity types for each domain. 

To make the popularity score fair, we filtered out entities that are likely too new to have sufficient statistical data for popularity measurement. For IMDb, MAG, DBLP, and Goodreads, we kept only entities by the year of 2020, 2020, 2020, and 2015, respectively. The cut-off years are all before the cut-off time of the LLM training data, so the benchmark avoids questions that require {\em recent} knowledge. We did not perform similar filtering for DBpedia because the year attribute is unavailable or non-applicable for most entities, and our pilot study shows that very few (if at all) of the questions generated from DBpedia require knowledge after 2020.

Table~\ref{tab:entitypartition} summarizes the distribution of head, torso, and tail entities. The distribution follows the {\em power law}, where very small percentages of entities fall in the head and torso buckets, and the majority of entities fall in the tail bucket; for example, over 99.9\% of movies fall in the tail bucket, according to IMDb vote counts. We also observe that this phenomenon is more pronounced when we measure by traffic than by density; for the latter, the torso buckets are often larger ($\sim$15\% of entities), and the tails are slightly smaller ($\sim$82\%).

\smallskip
\par{\noindent\textbf{Questions.}} 
We generated questions using a template-based approach, where each generated question asks for an attribute of an entity. We filtered out the following types of attributes: (i) unspecific (\eg, \texttt{seeAlso} in DBpedia), (ii) dynamic (\eg, \texttt{lastLaunchRocket} in DBpedia), (iii) data source specific (\eg, \texttt{averageRating} in IMDb), and (iv) non-textual (\eg, \texttt{picture} in DBpedia). We discuss a stricter filtering criterion in Appendix~\ref{sec:stricterfiltering}. %
For each specific domain (Movie, Book, Academics), we manually designed the question template for each attribute. %
DBpedia contains a large set of attributes, so we first employed ChatGPT to draft the templates (using Prompt~\ref{prompt:templatedraft} in Appendix~\ref{sec:prompt}), then proofread them manually and made necessary edits. Each question template corresponds to a distinct predicate. %

The answer for each question is the object of the relevant triple; when there are multiple answers (\eg, a book may have multiple authors), we included all in the answer. %
When necessary, we included extra information for an entity to avoid potential ambiguities (\eg, we included the publication year for a book to distinguish books of highly similar names). 

We generated an equal number of questions for randomly sampled head, torso, and tail entities using each template. For each specific domain, we generated $\sim$1K questions for each of the head, torso, and tail buckets. As DBPedia contains more domains and relationship types, we generated $\sim$3K questions for each bucket.
Table~\ref{tab:stats} summarizes the overall statistics of {\dn} in the number of questions and templates. %

\todo{add additional stats and split into subsections} %

\begin{table}[t!]
\centering
\scriptsize
\begin{tabular}{llrr}
\toprule
\bf Domain & \bf Sources & \bf \# Templates & \bf \# Questions \\
\midrule

Movie & IMDb & 13 & 3,093 \\
Book & Goodreads & 4 & 3,000 \\
Academics & MAG, DBLP & 13 & 2,946 \\
\noalign{\vskip 0.5ex}\hdashline\noalign{\vskip 0.5ex}
Open & DBpedia & 393 & 9,132 \\
\midrule
\bf Total &  & 423 & 18,171 \\

\bottomrule
\end{tabular}
\caption{The overall statistics of {\dn}.}
\label{tab:stats}
\end{table}

\subsection{Metrics}
\label{sec:metrics}
\par{\noindent\textbf{Metrics.}} We find that oftentimes LLMs are intelligent enough to admit that it does not have enough information to answer a question. As such, we used three metrics: {\em accuracy} ($\text{\bf A}$), {\em hallucination rate} ($\text{\bf H}$), and {\em missing rate} ($\text{\bf M}$), measuring the percentage of questions that an LLM gives the correct answer, gives a wrong or partially incorrect answer, or admits it cannot answer, respectively; by definition, $\text{A}+\text{H}+\text{M}=100\%$. 

Manually deciding the correctness of answers can be cumbersome. We next describe a few different ways to automatically decide if an answer is correct.

\smallskip
\par{\noindent\textbf{LLM-based.}} We ask ChatGPT to check whether an answer is correct given the question and ground truth (Prompt~\ref{prompt:answercheck} in Appendix~\ref{sec:prompt}). Thus, accuracy \textbf{A\textsubscript{LM}} is defined as the percentage of answers that ChatGPT judges as correct; hallucination rate \textbf{H\textsubscript{LM}} is defined as the percentage of time when (i) an attempted answer is not missing, and (ii) ChatGPT judges the answer as incorrect (\ie, $\text{H\textsubscript{LM}} = 100\% - \text{A\textsubscript{LM}} - \text{M}$). 

To understand the reliability of the LLM-based metrics, we randomly sampled $840$ answers from the evaluated LLMs and manually checked whether human judgment agrees with the LLM-based metrics. The agreement is $98\%$, which we view as reliable. Hence, we use A\textsubscript{LM} and H\textsubscript{LM} as the primary metrics in this study.

\smallskip
\par{\noindent\textbf{Rule-based.}} In addition, we adopt popular metrics, including {\em exact match (EM)}, {\em token F1 (F1)}, and {\em ROUGE-L (RL)}~\cite{lin2004rouge,rajpurkar2016squad}; in other words, we use rule-based methods to judge the correctness of an answer. Specifically, \textbf{A\textsubscript{EM}} is computed as the percentage of answers that exactly match the ground truth; \textbf{A\textsubscript{F1}} is computed as the average harmonic mean of precision and recall when comparing tokens in the returned answers and in the ground truth answers; \textbf{A\textsubscript{RL}} is computed as the average normalized longest common subsequence (LCS) between the returned answers and the ground truths. 
For common answer types, we additionally expand the set of ground-truth answers with their variants using hand-crafted rules (\eg, ``\textit{W Shakespeare}'' is a variant of ``\textit{William Shakespeare}''); when a given question has multiple expanded ground-truth answers, we take the maximum score. %

Correspondingly, we measure hallucination rate by \textbf{H\textsubscript{EM}} ($=100\%-\text{A\textsubscript{EM}}-\text{M}$), \textbf{H\textsubscript{F1}} ($=100\%-\text{A\textsubscript{F1}}-\text{M}$), and \textbf{H\textsubscript{RL}} ($=100\%-\text{A\textsubscript{RL}}-\text{M}$).
 As we will show later in Section~\ref{sec:rq4}, we observe high correlations between rule-based and LLM-based metrics. %

\subsection{Evaluation methodology}
\label{sec:evalmethod}
We prompted the LLM as shown in Prompt~\ref{prompt:original} in Appendix~\ref{sec:prompt}. First, we asked LLMs to give as concise answers as possible. Second, we prompted LLMs to respond ``unsure'' when the LLM is not confident in the answer. We applied few-shot learning and included in the prompt two examples that are not in {\dn}: one is a simple, answerable question with the corresponding answer as the response; the other is an unanswerable question with ``unsure'' as the response. 

With this prompt, rule-based metrics are more likely to reflect the factual correctness of the answers, and we can simply compute the missing rate (\ie, M) by counting ``unsure'' or empty answers. We observed that explicitly asking for ``unsure'' as an answer could significantly reduce hallucination rate (Section~\ref{sec:rq4}). %

To summarize, the following three setups in the benchmark and evaluation methodology help us best approximate the existence of (confident) knowledge in the LLMs: (i) focusing on simple questions in easy-to-understand forms, (ii) asking for concise answers to ease evaluation, and (iii) hinting the LLMs to answer ``unsure'' to suppress unnecessary hallucinations.

\section{Experimental Analysis}
\label{sec:experiment}

\begin{table*}[t!]
\centering
\scriptsize
\begin{tabular}{lrr|rr:rrrrrr}

\toprule
\multirow{2}{*}{\bf Model} & \multicolumn{2}{c|}{\textbf{All}} & \multicolumn{2}{c:}{\textbf{Open}} & \multicolumn{2}{c}{\textbf{Movie}} & \multicolumn{2}{c}{\textbf{Book}} & \multicolumn{2}{c}{\textbf{Academics}} \\

\cmidrule(rl){2-3} \cmidrule(rl){4-5} \cmidrule(rl){6-7} \cmidrule(rl){8-9} \cmidrule(rl){10-11} 

& \bf A\textsubscript{LM} & \bf H\textsubscript{LM} & \bf A\textsubscript{LM} & \bf H\textsubscript{LM} & \bf A\textsubscript{LM} & \bf H\textsubscript{LM} & \bf A\textsubscript{LM} & \bf H\textsubscript{LM} & \bf A\textsubscript{LM} & \bf H\textsubscript{LM} \\
\midrule

GPT-4 & 30.9 & 19.7 & 37.1 & 25.3 & 41.7 & 15.5 & 21.3 & 19.4 & 10.0 & 6.8 \\
ChatGPT & 20.3 & 14.1 & 22.1 & 14.8 & 34.7 & 13.3 & 16.9 & 24.9 & 3.0 & 1.9 \\
Llama 2 (70B) & 11.8 & 34.0 & 7.5 & 24.8 & 27.9 & 34.3 & 10.3 & 54.5 & 9.8 & 41.0 \\
LLaMA (33B) & 18.2 & 80.0 & 19.0 & 79.1 & 28.7 & 70.1 & 15.8 & 82.9 & 7.1 & 90.3 \\

\bottomrule
\end{tabular}

\caption{The best overall accuracy is only $\sim$31\% on {\dn}. All numbers are in percentage~(\%).}
\label{tab:rq1}
\end{table*}

\subsection{Models and configurations}
\label{sec:model}

We evaluated representative state-of-the-art LLMs of various sizes and architectures, including ChatGPT, GPT-4~\cite{openai2023gpt4}, LLaMA (7B, 13B, 33B, 65B)~\cite{touvron2023llama}, Llama~2 (70B)~\cite{touvron2023llama2}, Vicuna (7B, 13B)~\cite{chiang2023vicuna}, Flan-T5 (3B, 11B)~\cite{chung2022scaling}, RWKV (7B)~\cite{peng2023rwkv}, Falcon (7B, 40B), and Falcon-Instruct (7B, 40B)~\cite{almazrouei2023falcon}. %
We employed the most deterministic settings (\ie, \texttt{temperature=0} or \texttt{top\_k=1}) for all models. We present more details
in Appendix~\ref{sec:impldetails}.

Table~\ref{tab:mainresults} in Appendix~\ref{sec:supplemental} gives detailed results of all LLMs. We note that our goal is NOT to compare different LLM models; rather, by examining the metrics by different LLMs, we make sure to report the common patterns among the representative LLMs. We also note that it is hard to exhaustively benchmark every recent model in this fast-moving field; we conducted evaluations up to GPT-4~\cite{openai2023gpt4} and Llama~2~\cite{touvron2023llama2}, and detailed discussions can be based on slightly older models, where we observe similar patterns.

\subsection{RQ1: How reliable are LLMs in answering factual questions?}
\label{sec:rq1}

We present in Table~\ref{tab:rq1} the overall performance of GPT-4, ChatGPT, Llama 2-70B, and LLaMA-33B, which perform the best in most metrics on {\dn} among all LLMs introduced in Section~\ref{sec:model}. The best overall accuracy is obtained by GPT-4 at 31\%. %

Interestingly, for questions that are not answered correctly, different LLMs show different patterns: GPT-4 and ChatGPT give unsure or empty answers for the majority of them, and the hallucination rate is $<$20\% (still non-negligible); LLaMA-33B mostly provides hallucinated answers, resulting with high hallucination rate ($\sim$80\%); Llama 2-70B falls in-between. We suspect fine-tuning and reinforcement learning of these models may explain the different patterns when the model is unsure of the answers. Figure~\ref{tab:overview} %
shows examples of counterfactual answers given by GPT-4.

Finally, for all models, the overall performance varies substantially across different specific domains. All models perform the best in the {\em Movie} domain and worst in the {\em Academics} domain, likely because of the relatively low popularity of the {\em Academics} domain, as we will discuss soon.

\begin{table}[t!]
\centering
\scriptsize

\begin{subtable}[t]{0.48\textwidth}
\centering
\begin{tabular}{lrrrrrr}
\toprule
\multirow{2}{*}{\bf Domain} & \multicolumn{2}{c}{\textbf{Head}} & \multicolumn{2}{c}{\textbf{Torso}} & \multicolumn{2}{c}{\textbf{Tail}} \\
\cmidrule(rl){2-3} \cmidrule(rl){4-5} \cmidrule(rl){6-7}
& \bf A\textsubscript{LM} & \bf H\textsubscript{LM} & \bf A\textsubscript{LM} & \bf H\textsubscript{LM} & \bf A\textsubscript{LM} & \bf H\textsubscript{LM}  \\
\midrule

Movie & 59.3 & 14.8  & 55.0 & 16.9  & 10.9 & 14.7 \\
Book & 22.8 & 24.4  & 24.3 & 21.8  & 16.9 & 12.0 \\
Academics & 15.8 & 9.9  & 10.5 & 6.8  & 3.9 & 3.7 \\
\noalign{\vskip 0.5ex}\hdashline\noalign{\vskip 0.5ex}
Open & 47.6 & 30.2  & 36.5 & 24.1  & 27.3 & 21.6 \\
\midrule
All & 40.3 & 23.3  & 33.4 & 19.7  & 19.0 & 15.9 \\

\bottomrule
\end{tabular}
\caption{GPT-4.}
\end{subtable}

\begin{subtable}[t]{0.48\textwidth}
\centering
\begin{tabular}{lrrrrrr}
\toprule
\multirow{2}{*}{\bf Domain} & \multicolumn{2}{c}{\textbf{Head}} & \multicolumn{2}{c}{\textbf{Torso}} & \multicolumn{2}{c}{\textbf{Tail}} \\
\cmidrule(rl){2-3} \cmidrule(rl){4-5} \cmidrule(rl){6-7}
& \bf A\textsubscript{LM} & \bf H\textsubscript{LM} & \bf A\textsubscript{LM} & \bf H\textsubscript{LM} & \bf A\textsubscript{LM} & \bf H\textsubscript{LM}  \\
\midrule

Movie & 39.2 & 28.2  & 33.9 & 29.8  & 10.7 & 44.9 \\
Book & 15.0 & 52.2  & 12.9 & 54.4  & 3.1 & 56.9 \\
Academics & 12.9 & 35.2  & 11.1 & 38.2  & 5.3 & 49.7 \\
\noalign{\vskip 0.5ex}\hdashline\noalign{\vskip 0.5ex}
Open & 9.9 & 22.3  & 6.9 & 25.4  & 5.7 & 26.8 \\
\midrule
All & 16.2 & 30.3  & 13.2 & 33.0  & 6.1 & 38.6 \\

\bottomrule
\end{tabular}
\caption{Llama 2-70B.}
\end{subtable}

\caption{LLMs' factuality, measured by A\textsubscript{LM} (\%), decreases in the order of head, torso, and tail entities from {\dn}. }
\label{tab:rq2}
\end{table}

\begin{table}[t!]
\centering
\scriptsize

\begin{tabular}{lrrr}
\toprule
\bf Model & \bf A\textsubscript{LM} & \bf H\textsubscript{LM} & \bf M \\

\midrule
GPT-4 & 46.0 ($\uparrow$5.7) & 21.4 ($\downarrow$1.9) & 32.6 ($\downarrow$3.7) \\
Llama 2 (70B) & 18.7 ($\uparrow$2.5) & 29.7 ($\downarrow$0.6) & 51.6 ($\downarrow$1.9) \\

\bottomrule
\end{tabular}

\caption{Accuracy %
on the top-$10\%$ popular questions in the head bucket is only slightly better than overall head entities. ($\uparrow$/$\downarrow$: increased/decreased $\%$ compared with using all head instances.) } %
\label{tab:topofhead}

\end{table}

\subsection{RQ2: Do LLMs perform equally well on head, torso, and tail facts?}
\label{sec:rq2}

The overall accuracy of GPT-4 and Llama 2-70B (A\textsubscript{LM}) declines 
in the order of head, torso, and tail entities, as shown in Figure~\ref{tab:overview} and Table~\ref{tab:rq2}. We observe the same pattern for other LLMs. This verifies our hypothesis that as we lack training data for long-tail entities, it is difficult for LLMs to obtain knowledge for such entities. %

Surprisingly, the QA accuracy is still low even for the head entities (\eg, GPT-4 achieves an A\textsubscript{LM} of $48\%$ in the open domain). %
We further retain top-10\% popular questions from the head bucket. 
As shown in Table~\ref{tab:topofhead}, GPT-4 and Llama 2-70B obtained slightly higher accuracy (within 6 percent point) and lower hallucination rate for these super popular entities, but the accuracy is still disappointingly low (46\% for GPT-4 and 19\% for Llama 2-70B), and the missing rate is notable. We have a further discussion in Appendix~\ref{sec:missingratediscussion}. %

The QA accuracy on tail entities is significantly lower in most of the domains. Notably, {\em Academics} intuitively is a long-tail domain, and we observe $\sim$10\% overall accuracy and very low accuracy (16\% for GPT-4 and 13\% for Llama 2-70B) even for head entities in this domain. 

Finally, hallucination rate drops from head to torso to tail for GPT-4, but increases for Llama 2-70B. We hypothesize that there is at least one more factor that affects the hallucination rate---the internal assessment of the confidence. When an LLM ``knows'' what is unknown to it, it is likely to reduce confidence when answering related questions and produce fewer hallucinations.

\begin{table}[t!]
\centering
\scriptsize
\begin{tabular}{lrrrr}
\toprule
\multirow{2}{*}{\bf Model} & \multicolumn{2}{c}{\textbf{Head \& Torso}} & \multicolumn{2}{c}{\textbf{Tail}} \\
\cmidrule(rl){2-3} \cmidrule(rl){4-5}
& \bf A\textsubscript{LM} & \bf H\textsubscript{LM} & \bf A\textsubscript{LM} & \bf H\textsubscript{LM} \\
\midrule

GPT-4 & 42.9 & 20.3 & 36.8 & 25.6 \\
ChatGPT & 18.6 & 14.2 & 22.3 & 14.8 \\
Llama 2 (70B) & 8.0 & 42.1 & 7.5 & 23.8 \\
LLaMA (7B) & 15.3 & 83.5 & 13.5 & 77.7 \\
LLaMA (13B) & 14.6 & 85.1 & 14.7 & 83.6 \\
LLaMA (33B) & 18.2 & 81.4 & 19.0 & 78.9 \\
LLaMA (65B) & 20.1 & 79.7 & 18.3 & 81.4 \\
Vicuna (7B) & 12.5 & 82.0 & 9.3 & 77.4 \\
Vicuna (13B) & 13.0 & 70.3 & 8.6 & 55.5 \\
Flan-T5 (3B) & 4.4 & 13.0 & 3.4 & 10.4 \\
Flan-T5 (11B) & 9.2 & 11.1 & 5.0 & 8.1 \\
RWKV (7B) & 6.9 & 28.7 & 6.4 & 29.7 \\
Falcon (7B) & 11.3 & 51.0 & 8.1 & 43.7 \\
Falcon (40B) & 14.4 & 34.1 & 8.5 & 29.2 \\
Falcon-Instruct (7B) & 8.8 & 48.3 & 7.2 & 47.1 \\
Falcon-Instruct (40B) & 12.8 & 15.3 & 7.9 & 15.2 \\

\bottomrule
\end{tabular}
\caption{Comparison of LLMs' factuality about head, torso, and tail predicates in A\textsubscript{LM} (\%) and H\textsubscript{LM} (\%) using open-domain instances from {\dn}.}
\label{tab:headtotailattribute}

\end{table}

\smallskip
\par{\noindent\textbf{Head-to-tail predicates.}} We investigated whether the performance still correlates with the head-to-tail order regarding the popularity of \emph{predicates} instead of entities. We sorted the predicates from DBpedia by popularity (measured by the number of relational triples with the predicate) and partitioned the sorted predicates into head, torso, and tail in a similar fashion. We then re-partitioned the open-domain questions into head, torso, and tail predicate buckets, each containing $72$, $450$, and $8,610$ questions, respectively. Since the number of questions in the head bucket is low, we merged the head and torso buckets. 

Table~\ref{tab:headtotailattribute} compares the performance on head \& torso \vs on tail. 
We observe no consistent correlation among different LLMs between the performance and the head-to-tail predicate ordering, and the differences in accuracy are not very high. This is not too surprising for two reasons. First, the semantics of each predicate is mostly consistent with the semantics of the predicate names, which can be well understood by LLMs. Second, when facts are present for tail predicates, they are often about the head entities, and factual information for head entities is likely to be more abundant in the training data.

\begin{table*}[t!]
\centering
\scriptsize
\begin{tabular}{lrrrrrrrrr}
\toprule
\multirow{2}{*}{\bf Model} & \multicolumn{3}{c}{\textbf{Head-to-Tail}} & \multicolumn{2}{c}{\textbf{Head}} & \multicolumn{2}{c}{\textbf{Torso}} & \multicolumn{2}{c}{\textbf{Tail}} \\
\cmidrule(rl){2-4} \cmidrule(rl){5-6} \cmidrule(rl){7-8} \cmidrule(rl){9-10}
 & \bf A\textsubscript{LM} & \bf H\textsubscript{LM} & \bf M & \bf A\textsubscript{LM} & \bf H\textsubscript{LM} & \bf A\textsubscript{LM} & \bf H\textsubscript{LM} & \bf A\textsubscript{LM} & \bf H\textsubscript{LM} \\
\midrule

LLaMA (7B) & 12.1 & 80.0 & 7.9 & 19.0 & 74.4 & 11.7 & 81.0 & 5.4 & 84.8 \\
LLaMA (13B) & 14.4 & 84.3 & 1.3 & 22.0 & 77.2 & 14.8 & 83.8 & 6.3 & 91.9 \\
LLaMA (33B) & 18.2 & 80.0 & 1.8 & 26.0 & 72.8 & 19.8 & 78.7 & 8.8 & 88.6 \\
LLaMA (65B) & 17.8 & 81.9 & 0.3 & 25.9 & 73.8 & 18.7 & 81.0 & 8.7 & 90.9 \\
\midrule
Vicuna (7B) & 10.1 & 79.2 & 10.8 & 16.2 & 72.7 & 9.6 & 79.8 & 4.3 & 85.0 \\
Vicuna (13B) & 9.2 & 62.6 & 28.2 & 14.0 & 55.0 & 8.8 & 62.8 & 4.7 & 70.0 \\
\midrule
Flan-T5 (3B) & 2.3 & 17.4 & 80.3 & 3.9 & 19.7 & 1.5 & 17.1 & 1.3 & 15.5 \\
Flan-T5 (11B) & 4.2 & 20.0 & 75.7 & 7.6 & 23.7 & 3.2 & 19.9 & 2.0 & 16.5 \\
\midrule
Falcon (7B) & 9.5 & 57.9 & 32.6 & 14.5 & 53.8 & 9.2 & 57.9 & 4.8 & 62.0 \\
Falcon (40B) & 10.8 & 41.0 & 48.2 & 16.2 & 36.4 & 11.2 & 40.0 & 4.9 & 46.6 \\
\midrule
Falcon-Instruct (7B) & 6.8 & 56.7 & 36.5 & 11.5 & 56.0 & 5.6 & 57.2 & 3.4 & 56.7 \\
Falcon-Instruct (40B) & 10.8 & 32.2 & 57.0 & 16.7 & 30.5 & 11.5 & 31.1 & 4.3 & 34.8 \\

\bottomrule
\end{tabular}
\caption{Comparison of different LLMs with different sizes. All numbers are in percentage (\%).}
\label{tab:rq3}
\end{table*}

\subsection{RQ3: Does normal methods that improve LLMs increase the factuality?}
\label{sec:rq3}
Table~\ref{tab:rq3} compares LLMs in different sizes and with or without instruction tuning. First, we observe that an increased model size does not automatically translate to a better grasp of factual knowledge. For example, LLaMA-33B modestly outperforms LLaMA-65B across the head, torso, and tail subsets ($+0.4\%$ %
in A\textsubscript{LM} and $-1.9\%$ %
in H\textsubscript{LM} on average) while they share the same training dataset and hyperparameters. This provides additional evidence for our hypothesis that once the model is sufficiently large, the abundance of training data plays a more critical role in the factuality of the LLMs.

Second, compared with LLaMA and Falcon, the instruction-tuned counterparts (\ie, Vicuna and Falcon-Instruct) have lower accuracy, as they learned to be more conservative in providing factual answers and thus generate ``unsure'' more often (\eg, Vicuna-13B is $26.9\%$ %
higher in M than LLaMA-13B). Despite so, they still have high hallucination rate.

\begin{table}[t!]
\centering
\scriptsize
\begin{tabular}{llrrrr}
\toprule
\multirow{2}{*}{\bf LLM-Based} & \multirow{2}{*}{\bf Rule-Based} & \multicolumn{2}{c}{\textbf{$\rho$}} & \multicolumn{2}{c}{\textbf{$r$}} \\
& & Min. & Mean & Min. & Mean \\
\midrule

 & A\textsubscript{EM} & 0.721 & 0.915 & 0.921 & 0.966 \\
A\textsubscript{LM} & A\textsubscript{F1} & 0.775 & 0.951 & 0.781 & 0.969 \\
 & A\textsubscript{RL} & 0.730 & 0.947 & 0.775 & 0.969 \\
\midrule
 & H\textsubscript{EM} & 0.968 & 0.991 & 0.993 & 0.998 \\
H\textsubscript{LM} & H\textsubscript{F1} & 0.976 & 0.995 & 0.998 & 0.999 \\
 & H\textsubscript{RL} & 0.976 & 0.995 & 0.998 & 0.999 \\

\bottomrule
\end{tabular}
\caption{The minimum and mean Spearman's rank correlation coefficients ($\rho$) and Pearson correlation coefficients ($r$) show high correlation between LM- and rule-based metrics.} %
\label{tab:metriccorrelation}
\end{table}

\subsection{Robustness of our evaluation methodology}
\label{sec:rq4}

Finally, we evaluate the robustness of our evaluation methodology.

\smallskip
\par{\noindent\textbf{Correlations between rule- and LLM-based metrics.}} 
For each combination of popularity (head, torso, tail) and domain (movie, book, %
academics, open), we calculate Spearman's rank and Pearson correlation coefficients between rule- and LLM-based metrics over all LLMs. We report the aggregated results (minimum, mean) in Table~\ref{tab:metriccorrelation}. The correlation scores suggest that A\textsubscript{LM} (\resp H\textsubscript{LM}) strongly correlates with A\textsubscript{EM}, A\textsubscript{F1}, and A\textsubscript{RL} (\resp H\textsubscript{EM}, H\textsubscript{F1}, and H\textsubscript{RL}), indicating that rule-based metrics are good alternatives for lower-cost or faster evaluation.

\begin{table}[t!]
\centering
\scriptsize
\begin{tabular}{llrrrrrr}
\toprule
& \multirow{2}{*}{\bf Domain} & \multicolumn{2}{c}{\textbf{Few-shot}} & \multicolumn{2}{c}{\textbf{Zero-shot}} & \multicolumn{2}{c}{\textbf{In-domain}} \\
\cmidrule(rl){3-4} \cmidrule(rl){5-6} \cmidrule(rl){7-8}
& & \bf A\textsubscript{LM} & \bf H\textsubscript{LM} & \bf A\textsubscript{LM} & \bf H\textsubscript{LM} & \bf A\textsubscript{LM} & \bf H\textsubscript{LM} \\
\midrule

\multirow{2}{*}{\rotatebox[origin=c]{90}{\bf Head}} & Open & 32.7 & \bf 20.8 & 32.6 & 24.7 & \bf 45.0 & 27.8 \\
\cmidrule{2-8}
& All & 29.4 & \bf 17.2 & 29.2 & 18.6 & \bf 38.3 & 24.7 \\
\midrule
\multirow{2}{*}{\rotatebox[origin=c]{90}{\bf Torso}} & Open & 19.7 & \bf 13.3 & 21.6 & 17.9 & \bf 30.1 & 23.0 \\
\cmidrule{2-8}
& All & 21.9 & \bf 14.6 & 22.8 & 16.7 & \bf 29.8 & 22.8 \\
\midrule
\multirow{2}{*}{\rotatebox[origin=c]{90}{\bf Tail}} & Open & 13.8 & \bf 10.2 & 14.9 & 14.5 & \bf 23.0 & 19.5 \\
\cmidrule{2-8}
& All & 9.5 & \bf 10.5 & 10.3 & 12.7 & \bf 15.4 & 20.2 \\

\bottomrule
\end{tabular}
\caption{Performance of ChatGPT with different prompts on {\dn}. All numbers are in percentage (\%).} 
\label{tab:differentprompt}
\end{table}

\smallskip
\par{\noindent\textbf{Effect of brief and ``unsure''.}} 
We randomly sampled $1.2$K questions and tested the stability of answers if we call ChatGPT to regenerate answers. When not requiring brief or ``unsure'' answers, for $18\%$ of questions, ChatGPT regenerated different answers. Adding the requirement for brief answers (Prompt~\ref{prompt:variant3} in Appendix~\ref{sec:prompt}) reduced the percentage to $4\%$, and further asking ``unsure'' answers with few-shot examples (Prompt~\ref{prompt:original}) reduced the percentage to $1\%$. In addition, according to manual evaluation on $150$ randomly sampled questions, removing ``unsure'' as an option increases ChatGPT's hallucination rate by $13$ percentage points.

\smallskip
\par{\noindent\textbf{Robustness of prompts.}} 
We explore two other prompts. Compared with the original prompt that conducts few-shot learning (Section~\ref{sec:model}), denoted as \textbf{Few-shot}, the \textbf{Zero-shot} prompt does not provide examples and thus is zero-shot learning (Prompt~\ref{prompt:variant1} in Appendix~\ref{sec:prompt}), and the \textbf{In-domain} prompt has the answerable example swapped out for an in-domain example generated by the same question template as the target question (Prompt~\ref{prompt:variant2} in Appendix~\ref{sec:prompt}). 

As shown in Table~\ref{tab:differentprompt}, \textbf{Few-shot} and \textbf{Zero-shot} show very similar results, but performance differences are noticeable between \textbf{Few-shot} and \textbf{In-domain}. In particular, in-domain examples help get more correct answers ($+8.9\%$, $+7.9\%$, $+5.9\%$ %
in A\textsubscript{LM} for head, torso, tail) but at the cost of more hallucinations ($+7.5\%$, $+8.2\%$, $+9.7\%$ %
in H\textsubscript{LM} for head, torso, tail). We suspect that the in-domain examples boost the confidence of ChatGPT in answering a question, so it answers questions even when the real confidence is not that high, causing both higher accuracy and higher hallucination rate. %

Despite the fluctuation, our original prompt template (\textbf{Few-shot}) appears to be better at approximating the (confident) factuality of LLMs with the QA accuracy, and the 
\emph{relative} performance among the head, torso, and tail remains stable over different prompts.

\section{Discussions}
\label{sec:disc}
\subsection{The future of knowledge graphs}
The experimental analysis indicates that although LLMs have incorporated factual knowledge within their parameters, the amount of this encoded knowledge remains limited. Knowledge of long-tail entities is already sparse in KGs and is even more deficient in LLMs. 

Nevertheless, LLMs have been revolutionizing the way people seek information and calling for reconsideration of the best representation of factual knowledge. We term the forthcoming generation of KGs as {\em Dual Neural KGs}: knowledge can reside explicitly as triples (similar to KGs) and implicitly as embeddings (like in LLMs); the symbolic form caters to human understanding and explainability, while the neural form benefits machine comprehension and seamless conversations. A piece of knowledge can exist in both formats or in the one that is more appropriate. The harmonious blend of the two forms, capitalizing on the latest LLM innovations, is an exciting research area as we elaborate next.

\smallskip
\par{\noindent\textbf{Head knowledge.}} This involves popular entities where training data are ample. Ideally, LLMs could be taught such knowledge for efficient retrieval, meaning head knowledge shall exist in both forms. Currently, LLMs still have a mediocre QA accuracy for popular entities (see Table~\ref{tab:topofhead}), so a critical research area is to infuse head knowledge into LLMs through model training or fine-tuning. Early work in this line includes knowledge infusion~\cite{kg-bart, k-adapter, zhen2022survey}.

\smallskip
\par{\noindent\textbf{Torso-to-tail and recent knowledge.}} This involves non-popular entities and emerging knowledge, where training data are typically sparse or absent. This type of knowledge might be best represented as triples. Serving such knowledge requires effectively deciding when external knowledge is essential, efficiently retrieving the relevant knowledge, and seamlessly integrating it into the answers. Early attempts in this direction involve knowledge-augmented LLMs~\cite{asai2023retrieval, webgpt, replug, BMH+22}.

\subsection{Limitations and extensions}
\par{\noindent\textbf{Taxonomy.}} Our work does not discuss the effectiveness of LLMs in capturing taxonomy or type hierarchies, which could be an extension of this study. Specifically, we hypothesize that LLMs can effectively incorporate type relationships (\eg, hypernyms and synonyms), even for the fine-granularity sub-types. Hence, it may no longer be worth manually constructing a very deep and complex hierarchy in the future. %

\par{\noindent{\textbf{Robustness to question formulation.}}} This paper primarily aims to evaluate how much an LLM ``knows'' a fact with high confidence; we thus tested various ways of formulating factual questions and selected the least ambiguous form for this study. However, this approach does not assess the model's robustness to paraphrasing or consider the diverse ways models can be queried, such as entailment or cloze-style prompts. Our supplementary experiment in Appendix~\ref{sec:clozequery} suggests that varying the form of questions does not significantly impact the evaluation results. A more thorough evaluation of robustness is beyond the scope of this paper and left for future research.

\section{Related Work}

\par{\noindent\textbf{Benchmarks.}} Most works studied the factuality of LLMs using existing QA benchmarks such as WebQuestions~\cite{berant2013semantic}, TriviaQA~\cite{joshi2017triviaqa}, LC-QuAD~\cite{trivedi2017lcquad,dubey2019lcquad}, QALD-9~\cite{usbeck20189th}, Natural Questions~\cite{kwiatkowski2019natural}, and EntityQuestions~\cite{sciavolino2021simple}. A recent line of work has been constructing new QA benchmarks to assess LLMs' factuality, especially for long-tail knowledge~\cite{mallen2023not,kim2023automatic}. Compared with these benchmarks, {\dn} is the first to specifically assess how well LLMs incorporate head, torso, and tail factual information. 

\smallskip
\par{\noindent\textbf{LLM Evaluation.}} Recent years have seen a proliferation of research on assessing the factuality of LLMs~\cite{roberts2020much,petroni2021kilt,shuster2021retrieval,mielke2022reducing,tan2023evaluation,hu2023empirical,peng2023check,omar2023chatgpt,kandpal2023large,mallen2023not,chen2023knowledge}. Most of these works focus on a single knowledge source, such as Freebase or Wikipedia, and they have yet to systematically perform the evaluation explicitly regarding head/torso/tail entities or attributes. 
One work close to ours is \citet{omar2023chatgpt}, which evaluated ChatGPT using facts collected from diverse knowledge sources; however, their evaluation was carried out manually on only $450$ QA instances. 

There are three works that also showed the correlation between the QA accuracy %
of language models and fact popularity~\cite{mallen2023not,kandpal2023large,kim2023automatic}. %
Our work, conducted in parallel, focuses on a different angle---how knowledgeable are LLMs? For this purpose, we systematically designed experimental methodology, including the definition of head, torso, and tail entities, the design of metrics, and the evaluation method. Our benchmark is comprehensive in containing different knowledge sources, different domains, and rich relations. Compared with these three works, we gave more quantified answers for research questions RQ1--RQ3.%

\section{Conclusion}

We introduce {\dn}, the first benchmark designed to assess the ability of LLMs to internalize head, torso, and tail facts. Alongside the dataset, we present a new evaluation methodology with appropriate metrics for automatically evaluating LLMs' factuality. Our evaluation shows that even the most advanced LLMs have notable limitations in representing factual knowledge, particularly for the torso and tail entities. Accordingly, we suggest new research areas to seamlessly blend knowledge in the symbolic form and neural form.

\section*{Acknowledgements}
We would like to thank the anonymous ARR reviewers and meta reviewer for their constructive and insightful feedback.

\bibliography{custom}
\bibliographystyle{acl_natbib}

\appendix

\clearpage
\section{Appendix}

\todo{(knowledge cutoff date)}

\todo{data cleaning}

\subsection{List of Prompts}
\label{sec:prompt}

\begin{prompt}[ht!]
\centering
\scriptsize
\begin{tabular}{p{0.45\textwidth}}
\toprule
You are given a few samples of a relation in the format of <X, relation, Y>. You need to write a question *template* about the relation, which can be used to generate questions. The template needs to have one blank such that a question about Y can be generated by filling the blank with X.\\\\

\#Example 1\\
Samples: <!Hero, musicBy, Eddie DeGarmo>, <9 to 5 (musical), musicBy, Dolly Parton>, <All About Us (musical), musicBy, John Kander>\\
Template: The music of \_ is by whom?\\\\

\#Example 2\\
Samples: <10,000 Maniacs, bandMember, Dennis Drew>, <16bit (band), bandMember, Eddie Jefferys>, <1TYM, bandMember, Teddy Park>\\
Template: Name a band member of \_?\\\\

\#Example 3\\
Samples: \{SAMPLES\}\\
Template: \\
\bottomrule
\end{tabular}
\caption{Question template drafting.}
\label{prompt:templatedraft}

\end{prompt}

\begin{prompt}[ht!]
\centering
\scriptsize
\begin{tabular}{p{0.45\textwidth}}
\toprule
You need to check whether the prediction of a question-answering system to a question is correct. You should make the judgment based on a list of ground truth answers provided to you. Your response should be "correct" if the prediction is correct or "incorrect" if the prediction is wrong.\\\\

Question: Who authored The Taming of the Shrew (published in 2002)?\\
Ground truth: ["William Shakespeare", "Roma Gill"]\\
Prediction: W Shakespeare\\
Correctness: correct\\\\

Question: Who authored The Taming of the Shrew (published in 2002)?\\
Ground truth: ["William Shakespeare", "Roma Gill"]\\
Prediction: Roma Gill and W Shakespeare\\
Correctness: correct\\\\

Question: Who authored The Taming of the Shrew (published in 2002)?\\
Ground truth: ["William Shakespeare", "Roma Gill"]\\
Prediction: Roma Shakespeare\\
Correctness: incorrect\\\\

Question: What country is Maharashtra Metro Rail Corporation Limited located in?\\
Ground truth: ["India"]\\
Prediction: Maharashtra\\
Correctness: incorrect\\\\

Question: What's the job of Song Kang-ho in Parasite (2019)?\\
Ground truth: ["actor"]\\
Prediction: He plays the role of Kim Ki-taek, the patriarch of the Kim family.\\
Correctness: correct\\\\

Question: Which era did Michael Oakeshott belong to?\\
Ground truth: ["20th-century philosophy"]\\
Prediction: 20th century.\\
Correctness: correct\\\\

Question: Edward Tise (known for Full Metal Jacket (1987)) is in what department?\\
Ground truth: ["sound department"]\\
Prediction: 2nd Infantry Division, United States Army\\
Correctness: incorrect\\\\

Question: What wine region is Finger Lakes AVA a part of?\\
Ground truth: ["New York wine"]\\
Prediction: Finger Lakes AVA\\
Correctness: incorrect\\\\

Question: \{QUESTION\}\\
Ground truth: \{GROUND\_TRUTH\}\\
Prediction: \{PREDICTION\}\\
Correctness: \\
\bottomrule
\end{tabular}
\caption{Correctness checking.}
\label{prompt:answercheck}
\end{prompt}

\begin{prompt}[ht!]
\centering
\scriptsize
\begin{tabular}{p{0.45\textwidth}}
\toprule
Answer the following questions in as few words as possible. Say "unsure" if you don't know.\\\\

Question: What is the capital of China?\\
Answer: Beijing\\\\

Question: What is the captical of Wernythedia?\\
Answer: unsure\\\\

Question: \{QUESTION\}\\
Answer: \\
\bottomrule
\end{tabular}
\caption{Question answering (Few-shot).}
\label{prompt:original}
\end{prompt}

\begin{prompt}[ht!]
\centering
\scriptsize
\begin{tabular}{p{0.45\textwidth}}
\toprule
Answer the following question in as few words as possible. Say "unsure" if you don't know. \{QUESTION\} \\
\bottomrule
\end{tabular}
\caption{Question answering (Zero-shot).}
\label{prompt:variant1}
\end{prompt}

\begin{prompt}[ht!]
\centering
\scriptsize
\begin{tabular}{p{0.45\textwidth}}
\toprule

Answer the following questions in as few words as possible. Say "unsure" if you don't know.\\\\

Question: What is the captical of Wernythedia?\\
Answer: unsure\\\\

Question: \{QUESTION\textsuperscript{\#}\}\\
Answer: \{ANSWER\textsuperscript{\#}\}\\\\

Question: \{QUESTION\}\\
Answer: \\
\bottomrule
\end{tabular}
\caption{Question answering (In-domain) (\#: the in-domain instance described in Section~\ref{sec:rq4}).}
\label{prompt:variant2}
\end{prompt}

\begin{prompt}[ht!]
\centering
\scriptsize
\begin{tabular}{p{0.45\textwidth}}
\toprule

Answer the following questions in as few words as possible. \{QUESTION\} \\
\bottomrule
\end{tabular}
\caption{Question answering (simply asking for concise answers).}
\label{prompt:variant3}
\end{prompt}

\begin{prompt}[ht!]
\centering
\scriptsize
\begin{tabular}{p{0.45\textwidth}}
\toprule
Answer the following questions in as few words as possible. Return your best guess if you don't know.\\\\

Question: What is the capital of China?\\
Answer: Beijing\\\\

Question: \{QUESTION\}\\
Answer: \\
\bottomrule
\end{tabular}
\caption{Question answering (returning its best guess instead of ``unsure'' when the confidence is low).}
\label{prompt:bestguess}
\end{prompt}

\subsection{Popularity measure in head-to-tail partition}
\label{sec:popularitymeasure}

\begin{itemize}
    \item \textbf{IMDb} (traffic): The number of votes (\ie, \texttt{numVotes}) the {\em title} (\eg, movie, short, TV series, etc.) has received; we do NOT consider whether the vote is high or low in the counting. For person entities, we use the total number of votes received by the titles the person is known for.
    \item \textbf{Goodreads} (traffic): The count of ratings (\ie, \texttt{ratings\_count}) the book has received; similarly, we do NOT take into consideration whether the rating is high or low.
    \item \textbf{MAG} (traffic): The number of citations (\ie, \texttt{CitationCount}) the entity (\ie, scholarly article, conference, or journal) has received. 
    \item \textbf{DBLP} (density): The number of works the scholar has authored.
    \item \textbf{DBpedia} (density): The number of relational triples in DBPedia that contain the entity.
\end{itemize}

\subsection{Implementation details}
\label{sec:impldetails}

We interacted with ChatGPT and GPT-4 through OpenAI API\footnote{\url{https://platform.openai.com/docs/api-reference}}. The employed version of ChatGPT and GPT-4 is \texttt{gpt-3.5-turbo-0301} and \texttt{gpt-4-0613}, respectively. We used Transformers~\cite{wolf2020transformers} to interact with the other LLMs on A100 (80GB) GPUs, and we used 16-bit floating point formats (\ie, float16 for Flan-T5 and RWKV, bfloat16 for LLaMA, Llama 2, Vicuna, Falcon, and Falcon-Instruct). We employed the original LLaMA, Llama 2, Flan-T5, Falcon, and Falcon-Instruct versions. The employed version of RWKV and Vicuna is \texttt{v4 Raven} and \texttt{v1.1}, respectively. 

\subsection{Impact of less naturally occurring questions}
\label{sec:stricterfiltering}

\begin{table}[h!]
\centering
\scriptsize
\setlength{\tabcolsep}{5pt}
\begin{tabular}{lrrrrrr}

\toprule
\multirow{2}{*}{\bf Model} & \multicolumn{2}{c}{\textbf{Movie}} & \multicolumn{2}{c}{\textbf{Book}} & \multicolumn{2}{c}{\textbf{Academics}} \\

\cmidrule(rl){2-3} \cmidrule(rl){4-5} \cmidrule(rl){6-7} 

& \bf A\textsubscript{LM} & \bf H\textsubscript{LM} & \bf A\textsubscript{LM} & \bf H\textsubscript{LM} & \bf A\textsubscript{LM} & \bf H\textsubscript{LM} \\
\midrule

GPT-4 & 43.8 & 12.6 & 39.1 & 24.6 & 11.2 & 9.8 \\
ChatGPT & 37.8 & 14.5 & 31.0 & 21.5 & 2.3 & 1.6 \\
Llama 2 (70B) & 30.4 & 31.6 & 19.7 & 10.0 & 4.5 & 59.0 \\
LLaMA (7B) & 18.7 & 71.9 & 21.4 & 59.1 & 2.4 & 94.9 \\
LLaMA (13B) & 25.3 & 73.2 & 24.4 & 74.5 & 4.6 & 93.7 \\
LLaMA (33B) & 31.2 & 67.5 & 31.7 & 65.9 & 5.2 & 91.7 \\
LLaMA (65B) & 27.1 & 72.4 & 32.3 & 66.5 & 8.2 & 91.8 \\
Vicuna (7B) & 21.1 & 68.5 & 19.2 & 62.9 & 2.6 & 91.3 \\
Vicuna (13B) & 20.3 & 58.5 & 11.4 & 38.7 & 3.1 & 77.4 \\
Flan-T5 (3B) & 1.7 & 15.1 & 2.8 & 4.5 & 0.2 & 4.3 \\
Flan-T5 (11B) & 6.3 & 22.1 & 6.6 & 10.1 & 0.8 & 16.0 \\
RWKV (7B) & 4.5 & 24.3 & 13.3 & 37.0 & 0.1 & 9.5 \\
Falcon (7B) & 20.4 & 62.3 & 15.0 & 45.1 & 3.5 & 80.7 \\
Falcon (40B) & 26.0 & 43.1 & 11.4 & 7.1 & 4.7 & 55.0 \\
Falcon-Instruct (7B) & 13.4 & 66.4 & 9.9 & 34.1 & 1.9 & 58.5 \\
Falcon-Instruct (40B) & 28.4 & 37.5 & 11.9 & 1.9 & 3.9 & 51.2 \\

\bottomrule
\end{tabular}

\caption{Comparison of LLMs’ factuality on {\dn} without relatively less naturally occurring questions. All numbers are in percentage (\%).}
\label{tab:performanceofselectedattr}
\end{table}

\noindent When constructing {\dn}, we include all predicates that allow reasonable factual questions. Table~\ref{tab:performanceofselectedattr}, instead, shows metrics on predicates that users are more likely to ask about. In general we observed higher performance on the {\em Movie} and {\em Book} domains, but the accuracy is still fairly low and we observe similar patterns regarding head, torso, and tail entities.

\subsection{Asking questions in different forms}
\label{sec:clozequery}
\begin{table}[h!]
\centering
\scriptsize

\begin{tabular}{lrrrrrr}
\toprule
 & \multicolumn{2}{c}{\textbf{Head}} & \multicolumn{2}{c}{\textbf{Torso}} & \multicolumn{2}{c}{\textbf{Tail}} \\
\cmidrule(rl){2-3} \cmidrule(rl){4-5} \cmidrule(rl){6-7}
& \bf A\textsubscript{LM} & \bf H\textsubscript{LM} & \bf A\textsubscript{LM} & \bf H\textsubscript{LM} & \bf A\textsubscript{LM} & \bf H\textsubscript{LM}  \\
\midrule

Original & 51.3 & 11.5 & 46.4 & 16.6 & 6.4 & 11.8 \\
Cloze-style & 50.8 & 8.9 & 46.1 & 14.3 & 6.8 & 12.6 \\

\bottomrule
\end{tabular}

\caption{ChatGPT's factuality in A\textsubscript{LM} (\%) and H\textsubscript{LM} (\%) obtained by the cloze-style queries closely mirrors that of the simple-formed questions in the {\em Movie} domain.}
\label{tab:cloze}
\end{table}

\noindent We explored the influence of question formulation on the evaluation results using ChatGPT in the {\em Movie} domain. We rewrote all questions as cloze-style questions (\eg, ``What's the release year of Mr. \& Mrs. Smith'' was transformed to ``The release year of Mr. \& Mrs. Smith is \_''). As shown in Table~\ref{tab:cloze}, the performance obtained by the cloze-style queries is very similar to that obtained by simple-formed questions.

\subsection{Further discussions on the missing rate}
\label{sec:missingratediscussion}

\begin{table}[h!]
\centering
\scriptsize

\begin{tabular}{lrrr}
\toprule
\bf Domain & \bf A\textsubscript{LM} & \bf H\textsubscript{LM} & \bf M \\

\midrule
Movie & 63.5 & 13.5 & 22.9 \\
Academics & 25.3 & 14.7 & 60.0 \\
\bottomrule
\end{tabular}

\caption{Performance of GPT-4 on the top-$10\%$ popular questions in the head bucket. All numbers are in percentage (\%).} 
\label{tab:missingdisscussion1}

\end{table}

\begin{table}[h!]
\centering
\scriptsize

\begin{tabular}{lrrr}
\toprule
\bf Prompt & \bf A\textsubscript{LM} & \bf H\textsubscript{LM} & \bf M \\

\midrule
Original (Prompt~\ref{prompt:original}) & 63.5 & 13.5 & 22.9 \\
Prompt~\ref{prompt:bestguess} & 68.8 & 16.7 & 14.6 \\
\bottomrule
\end{tabular}

\caption{Performance of GPT-4 on the top-$10\%$ popular questions in the head bucket in the {\em Movie} domain. All numbers are in percentage (\%).} 
\label{tab:missingdisscussion2}

\end{table}

\noindent It is observed that even for the top-$10\%$ popular questions in the head bucket, the missing rate of GPT-4 is still over $30\%$ (Table~\ref{tab:topofhead}). Although this might seem counterintuitive, there are two reasons. First, the performance reported in Table~\ref{tab:topofhead} is based on all the studied domains, including the tail domain {\em Academics}. Table~\ref{tab:missingdisscussion1} compares GPT-4’s performance on the top-$10\%$ head entities in the {\em Academics} and the {\em Movie} domains. The missing rate on the more popular domain {\em Movie} is much lower ($23\%$). Second, if we explicitly ask the LLM to return the best guess (Prompt~\ref{prompt:bestguess}) instead of responding ``unsure'', GPT-4’s missing rate on the top $10\%$ of head entities in the {\em Movie} domain would further drop to $15\%$ (Table~\ref{tab:missingdisscussion2}). However, this is with the price of higher hallucination rate, showing that the confidence of this part of knowledge is low. Interestingly, even after the above change, GPT-4 still admits to being ``unsure'' for $15\%$ of questions (\eg, GPT-4’s answers are ``unknown'' given the questions ``What is the death year of Debbi Datz-Pyle (known for The Matrix (1999))?'', ``What movie is Alan R. Kessler known for?''). This further confirms that LLMs are not good at memorizing (internalizing) factual information.

\subsection{An example of entity bucketing}
\label{sec:entitybucketingexample}

\noindent Suppose there are $12$ entities $A,B,C,\ldots,L$, and their popularity scores are $A=8$, $B=4$, $C=D=2$, $E=F=\ldots=L=1$. The total popularity scores add up to $24$ ($=8+4+2+2+1\times8$). Top-$1/3$ traffic (a total score of $8$) is contributed by $\{A\}$, thus the head; mid-$1/3$ traffic is contributed by $\{B,C,D\}$ (a total score of $4+2+2=8$), thus the torso; bottom-$1/3$ traffic is contributed by $\{E,F,\ldots,L\}$ (a total score of $1\times8=8$), thus the tail. 

\clearpage

\subsection{Supplemental Results}
\label{sec:supplemental}

\begin{minipage}{\textwidth}
\centering
\tiny
\setlength{\tabcolsep}{5pt}
\begin{tabular}{llrrrrrrrrr|rrrrrr:rr}
\toprule
& \multirow{2}{*}{\bf Model} & \multicolumn{9}{c}{\textbf{All}} & \multicolumn{2}{|c}{\textbf{Movie}} & \multicolumn{2}{c}{\textbf{Book}} & \multicolumn{2}{c}{\textbf{Academics}} & \multicolumn{2}{:c}{\textbf{Open}} \\

\cmidrule(rl){3-11} \cmidrule(rl){12-13} \cmidrule(rl){14-15} \cmidrule(rl){16-17} \cmidrule(rl){18-19} 
&  & \bf A\textsubscript{EM} & \bf H\textsubscript{EM} & \bf A\textsubscript{F1} & \bf H\textsubscript{F1} & \bf A\textsubscript{RL} & \bf H\textsubscript{RL} & \bf A\textsubscript{LM} & \bf H\textsubscript{LM} & \bf M & \bf A\textsubscript{LM} & \bf H\textsubscript{LM} & \bf A\textsubscript{LM} & \bf H\textsubscript{LM} & \bf A\textsubscript{LM} & \bf H\textsubscript{LM} & \bf A\textsubscript{LM} & \bf H\textsubscript{LM} \\
\midrule

\multirow{16}{*}{\rotatebox[origin=c]{90}{\bf Head}} & GPT-4 & \bf 31.1 & 32.6 & \bf 37.2 & 26.5 & \bf 37.1 & 26.5 & \bf 40.3 & 23.3 & 36.3 & \bf 59.3 & 14.8 & 22.8 & \bf 24.4 & 15.8 & 9.9 & \bf 47.6 & 30.2 \\
& ChatGPT & 21.8 & 24.9 & 25.6 & 21.1 & 25.6 & 21.1 & 29.4 & \bf 17.2 & 53.3 & 51.3 & \bf 11.5 & 20.1 & 26.3 & 5.9 & \bf 3.0 & 32.7 & 20.8 \\
& Llama 2 (70B) & 13.9 & 32.7 & 16.5 & 30.0 & 16.5 & 30.1 & 16.2 & 30.3 & 53.5 & 39.2 & 28.2 & 15.0 & 52.2 & 12.9 & 35.2 & 9.9 & 22.3 \\
& LLaMA (7B) & 10.4 & 83.0 & 15.5 & 77.9 & 15.4 & 78.0 & 19.0 & 74.4 & 6.6 & 27.2 & 69.5 & 21.6 & 74.8 & 5.0 & 90.6 & 19.9 & 70.7 \\
& LLaMA (13B) & 12.7 & 86.6 & 18.1 & 81.1 & 18.0 & 81.2 & 22.0 & 77.2 & 0.8 & 36.5 & 63.1 & 20.1 & 79.9 & 9.7 & 89.8 & 21.7 & 77.1 \\
& LLaMA (33B) & 16.7 & 82.0 & 22.3 & 76.5 & 22.2 & 76.5 & 26.0 & 72.8 & 1.3 & 42.9 & 57.0 & \bf 24.2 & 75.8 & 10.8 & 87.5 & 25.8 & 72.3 \\
& LLaMA (65B) & 14.9 & 84.8 & 21.7 & 78.0 & 21.6 & 78.1 & 25.9 & 73.8 & 0.3 & 37.0 & 62.3 & 23.1 & 76.9 & \bf 16.5 & 83.5 & 26.1 & 73.7 \\
& Vicuna (7B) & 9.4 & 79.6 & 13.7 & 75.3 & 13.6 & 75.4 & 16.2 & 72.7 & 11.0 & 30.1 & 59.7 & 18.6 & 75.6 & 3.9 & 91.4 & 14.8 & 70.2 \\
& Vicuna (13B) & 8.7 & 60.3 & 11.9 & 57.1 & 11.9 & 57.1 & 14.0 & 55.0 & 31.0 & 29.9 & 52.0 & 10.8 & 64.0 & 5.4 & 74.3 & 12.5 & 46.8 \\
& Flan-T5 (3B) & 2.5 & \bf 21.1 & 3.3 & \bf 20.3 & 3.3 & \bf 20.3 & 3.9 & 19.7 & 76.4 & 2.3 & 19.9 & 3.7 & 52.1 & 0.1 & 7.9 & 5.7 & 12.8 \\
& Flan-T5 (11B) & 5.8 & 25.5 & 7.4 & 23.9 & 7.4 & 23.9 & 7.6 & 23.7 & 68.7 & 10.7 & 30.1 & 7.9 & 59.5 & 0.4 & 21.0 & 8.8 & \bf 10.6 \\
& RWKV (7B) & 6.2 & 35.3 & 8.3 & 33.2 & 8.2 & 33.2 & 9.6 & 31.9 & 58.5 & 9.8 & 26.7 & 15.4 & 49.8 & 0.2 & 13.2 & 10.7 & 33.7 \\
& Falcon (7B) & 10.0 & 58.3 & 12.9 & 55.4 & 12.8 & 55.5 & 14.5 & 53.8 & 31.7 & 28.1 & 51.6 & 14.8 & 68.3 & 8.1 & 80.4 & 11.9 & 41.1 \\
& Falcon (40B) & 12.1 & 40.5 & 14.5 & 38.1 & 14.6 & 38.0 & 16.2 & 36.4 & 47.4 & 36.0 & 30.3 & 10.2 & 52.9 & 11.5 & 61.8 & 13.1 & 24.8 \\
& Falcon-Instruct (7B) & 6.6 & 60.9 & 9.3 & 58.2 & 9.3 & 58.3 & 11.5 & 56.0 & 32.4 & 20.8 & 60.1 & 11.4 & 65.7 & 1.6 & 67.7 & 11.6 & 47.7 \\
& Falcon-Instruct (40B) & 12.4 & 34.8 & 15.2 & 32.0 & 15.2 & 32.0 & 16.7 & 30.5 & 52.7 & 39.4 & 27.6 & 9.8 & 51.0 & 10.9 & 60.0 & 13.2 & 15.3 \\
\midrule
\multirow{16}{*}{\rotatebox[origin=c]{90}{\bf Torso}} & GPT-4 & \bf 25.6 & 27.5 & \bf 30.9 & 22.3 & \bf 30.8 & 22.3 & \bf 33.4 & 19.7 & 46.9 & \bf 55.0 & 16.9 & \bf 24.3 & \bf 21.8 & 10.5 & 6.8 & \bf 36.5 & 24.1 \\
& ChatGPT & 16.6 & 20.0 & 19.0 & 17.5 & 19.0 & 17.5 & 21.9 & \bf 14.6 & 63.5 & 46.4 & 16.6 & 22.5 & 29.2 & 2.3 & \bf 1.7 & 19.7 & 13.3 \\
& Llama 2 (70B) & 11.4 & 34.8 & 13.7 & 32.5 & 13.6 & 32.5 & 13.2 & 33.0 & 53.8 & 33.9 & 29.8 & 12.9 & 54.4 & \bf 11.1 & 38.2 & 6.9 & 25.4 \\
& LLaMA (7B) & 5.7 & 87.0 & 9.7 & 83.0 & 9.6 & 83.1 & 11.7 & 81.0 & 7.3 & 21.2 & 72.3 & 9.5 & 80.4 & 2.6 & 93.7 & 12.2 & 80.0 \\
& LLaMA (13B) & 8.5 & 90.1 & 12.6 & 86.0 & 12.5 & 86.1 & 14.8 & 83.8 & 1.4 & 28.8 & 70.5 & 14.2 & 85.3 & 6.3 & 92.2 & 12.9 & 85.2 \\
& LLaMA (33B) & 12.7 & 85.7 & 17.5 & 80.9 & 17.5 & 81.0 & 19.8 & 78.7 & 1.5 & 36.0 & 63.8 & 19.2 & 80.5 & 7.9 & 88.4 & 18.4 & 79.9 \\
& LLaMA (65B) & 11.4 & 88.2 & 16.5 & 83.2 & 16.4 & 83.3 & 18.7 & 81.0 & 0.3 & 32.5 & 66.9 & 20.0 & 79.1 & 9.4 & 90.6 & 16.7 & 83.2 \\
& Vicuna (7B) & 5.4 & 84.0 & 8.7 & 80.7 & 8.6 & 80.8 & 9.6 & 79.8 & 10.6 & 22.9 & 64.6 & 8.7 & 82.1 & 2.6 & 91.6 & 7.7 & 80.3 \\
& Vicuna (13B) & 5.4 & 66.2 & 8.1 & 63.5 & 8.1 & 63.6 & 8.8 & 62.8 & 28.4 & 20.7 & 57.3 & 4.7 & 66.7 & 4.0 & 75.2 & 7.7 & 59.4 \\
& Flan-T5 (3B) & 0.7 & \bf 17.9 & 1.3 & \bf 17.3 & 1.3 & \bf 17.3 & 1.5 & 17.1 & 81.4 & 1.2 & \bf 14.2 & 0.5 & 51.4 & 0.1 & 7.1 & 2.5 & 10.0 \\
& Flan-T5 (11B) & 2.0 & 21.1 & 3.3 & 19.8 & 3.3 & 19.8 & 3.2 & 19.9 & 76.9 & 4.0 & 24.7 & 1.4 & 53.7 & 0.8 & 19.6 & 4.3 & \bf 7.3 \\
& RWKV (7B) & 2.0 & 26.9 & 3.4 & 25.5 & 3.3 & 25.5 & 3.8 & 25.0 & 71.1 & 2.3 & 24.4 & 4.1 & 33.8 & 0.1 & 7.6 & 5.4 & 28.0 \\
& Falcon (7B) & 5.9 & 61.3 & 8.3 & 58.8 & 8.3 & 58.9 & 9.2 & 57.9 & 32.9 & 20.5 & 54.6 & 6.5 & 74.0 & 6.2 & 83.4 & 7.3 & 45.6 \\
& Falcon (40B) & 8.4 & 42.8 & 10.1 & 41.1 & 10.1 & 41.1 & 11.2 & 40.0 & 48.9 & 28.0 & 34.6 & 5.8 & 52.1 & 9.3 & 62.7 & 7.9 & 30.5 \\
& Falcon-Instruct (7B) & 2.8 & 60.0 & 5.0 & 57.8 & 5.0 & 57.9 & 5.6 & 57.2 & 37.2 & 11.2 & 67.6 & 2.8 & 67.8 & 1.4 & 67.8 & 5.9 & 46.8 \\
& Falcon-Instruct (40B) & 8.7 & 33.9 & 11.0 & 31.6 & 10.9 & 31.7 & 11.5 & 31.1 & 57.4 & 31.7 & 31.2 & 7.1 & 50.4 & 9.4 & 61.5 & 6.7 & 14.9 \\
\midrule
\multirow{16}{*}{\rotatebox[origin=c]{90}{\bf Tail}} & GPT-4 & \bf 13.4 & 21.6 & \bf 17.1 & 17.8 & \bf 17.1 & 17.9 & \bf 19.0 & 15.9 & 65.1 & \bf 10.9 & 14.7 & \bf 16.9 & \bf 12.0 & 3.9 & 3.7 & \bf 27.3 & 21.6 \\
& ChatGPT & 5.9 & \bf 14.0 & 7.7 & \bf 12.2 & 7.7 & \bf 12.2 & 9.5 & \bf 10.5 & 80.1 & 6.4 & 11.8 & 8.0 & 19.2 & 0.8 & \bf 0.9 & 13.8 & 10.2 \\
& Llama 2 (70B) & 4.3 & 40.4 & 6.7 & 38.0 & 6.6 & 38.0 & 6.1 & 38.6 & 55.4 & 10.7 & 44.9 & 3.1 & 56.9 & \bf 5.3 & 49.7 & 5.7 & 26.8 \\
& LLaMA (7B) & 1.5 & 88.7 & 4.5 & 85.7 & 4.4 & 85.8 & 5.4 & 84.8 & 9.8 & 3.2 & 80.4 & 2.0 & 82.5 & 1.1 & 95.8 & 8.7 & 83.4 \\
& LLaMA (13B) & 1.9 & 96.3 & 5.0 & 93.1 & 5.0 & 93.2 & 6.3 & 91.9 & 1.8 & 4.8 & 92.1 & 2.4 & 96.5 & 1.9 & 96.6 & 9.5 & 88.7 \\
& LLaMA (33B) & 4.1 & 93.3 & 7.8 & 89.5 & 7.8 & 89.6 & 8.8 & 88.6 & 2.6 & 7.3 & 89.4 & 4.1 & 92.5 & 2.6 & 95.1 & 12.8 & 84.9 \\
& LLaMA (65B) & 3.5 & 96.2 & 7.4 & 92.2 & 7.4 & 92.2 & 8.7 & 90.9 & 0.4 & 5.4 & 94.4 & 5.9 & 93.2 & 3.3 & 96.6 & 12.5 & 87.1 \\
& Vicuna (7B) & 1.4 & 87.9 & 3.9 & 85.4 & 3.9 & 85.4 & 4.3 & 85.0 & 10.7 & 5.1 & 85.6 & 1.5 & 86.7 & 1.5 & 90.4 & 5.9 & 82.4 \\
& Vicuna (13B) & 1.6 & 73.1 & 3.9 & 70.8 & 3.9 & 70.8 & 4.7 & 70.0 & 25.3 & 5.7 & 72.6 & 1.7 & 77.2 & 1.6 & 81.8 & 6.4 & 62.8 \\
& Flan-T5 (3B) & 0.6 & 16.2 & 1.1 & 15.7 & 1.1 & 15.7 & 1.3 & 15.5 & 83.2 & 1.1 & \bf 8.0 & 0.2 & 53.0 & 0.4 & 6.0 & 2.1 & 8.8 \\
& Flan-T5 (11B) & 1.2 & 17.3 & 2.3 & 16.2 & 2.3 & 16.2 & 2.0 & 16.5 & 81.5 & 2.8 & 9.6 & 0.6 & 51.2 & 0.6 & 18.5 & 2.6 & \bf 6.8 \\
& RWKV (7B) & 0.5 & 22.1 & 1.6 & 21.1 & 1.6 & 21.1 & 1.8 & 20.9 & 77.4 & 0.3 & 16.4 & 0.4 & 16.5 & 0.0 & 10.3 & 3.3 & 27.2 \\
& Falcon (7B) & 2.1 & 64.6 & 4.2 & 62.6 & 4.2 & 62.6 & 4.8 & 62.0 & 33.2 & 7.6 & 71.2 & 1.4 & 75.2 & 1.9 & 89.6 & 5.8 & 45.7 \\
& Falcon (40B) & 2.7 & 48.8 & 4.3 & 47.2 & 4.3 & 47.2 & 4.9 & 46.6 & 48.5 & 7.6 & 54.2 & 1.3 & 55.5 & 3.7 & 71.3 & 5.6 & 33.2 \\
& Falcon-Instruct (7B) & 1.5 & 58.6 & 2.9 & 57.2 & 2.9 & 57.2 & 3.4 & 56.7 & 39.9 & 5.1 & 65.7 & 0.9 & 67.3 & 1.0 & 67.1 & 4.3 & 46.8 \\
& Falcon-Instruct (40B) & 2.3 & 36.8 & 4.3 & 34.9 & 4.2 & 34.9 & 4.3 & 34.8 & 60.9 & 7.0 & 44.6 & 1.0 & 51.3 & 3.9 & 67.7 & 4.6 & 15.5 \\
\midrule
\multirow{16}{*}{\rotatebox[origin=c]{90}{\bf Head-to-Tail}} & GPT-4 & \bf 23.3 & 27.2 & \bf 28.4 & 22.2 & \bf 28.3 & 22.2 & \bf 30.9 & 19.7 & 49.4 & \bf 41.7 & 15.5 & \bf 21.3 & \bf 19.4 & \bf 10.0 & 6.8 & \bf 37.1 & 25.3 \\
& ChatGPT & 14.7 & 19.6 & 17.4 & \bf 16.9 & 17.4 & \bf 16.9 & 20.3 & \bf 14.1 & 65.6 & 34.7 & \bf 13.3 & 16.9 & 24.9 & 3.0 & \bf 1.9 & 22.1 & 14.8 \\
& Llama 2 (70B) & 9.8 & 35.9 & 12.3 & 33.5 & 12.3 & 33.5 & 11.8 & 34.0 & 54.2 & 27.9 & 34.3 & 10.3 & 54.5 & 9.8 & 41.0 & 7.5 & 24.8 \\
& LLaMA (7B) & 5.9 & 86.2 & 9.9 & 82.2 & 9.8 & 82.3 & 12.1 & 80.0 & 7.9 & 17.2 & 74.1 & 11.0 & 79.2 & 2.9 & 93.4 & 13.6 & 78.0 \\
& LLaMA (13B) & 7.7 & 91.0 & 11.9 & 86.8 & 11.8 & 86.8 & 14.4 & 84.3 & 1.3 & 23.3 & 75.3 & 12.2 & 87.2 & 6.0 & 92.9 & 14.7 & 83.7 \\
& LLaMA (33B) & 11.2 & 87.0 & 15.9 & 82.3 & 15.8 & 82.4 & 18.2 & 80.0 & 1.8 & 28.7 & 70.1 & 15.8 & 82.9 & 7.1 & 90.3 & 19.0 & 79.1 \\
& LLaMA (65B) & 9.9 & 89.8 & 15.2 & 84.5 & 15.1 & 84.5 & 17.8 & 81.9 & 0.3 & 25.0 & 74.5 & 16.3 & 83.1 & 9.7 & 90.3 & 18.4 & 81.3 \\
& Vicuna (7B) & 5.4 & 83.8 & 8.8 & 80.5 & 8.7 & 80.5 & 10.1 & 79.2 & 10.8 & 19.4 & 70.0 & 9.6 & 81.5 & 2.7 & 91.2 & 9.5 & 77.6 \\
& Vicuna (13B) & 5.2 & 66.5 & 8.0 & 63.8 & 7.9 & 63.8 & 9.2 & 62.6 & 28.2 & 18.8 & 60.7 & 5.7 & 69.3 & 3.7 & 77.1 & 8.9 & 56.4 \\
& Flan-T5 (3B) & 1.3 & \bf 18.4 & 1.9 & 17.8 & 1.9 & 17.8 & 2.3 & 17.4 & 80.3 & 1.5 & 14.0 & 1.5 & 52.2 & 0.2 & 7.0 & 3.4 & 10.5 \\
& Flan-T5 (11B) & 3.0 & 21.3 & 4.3 & 20.0 & 4.3 & 20.0 & 4.2 & 20.0 & 75.7 & 5.8 & 21.5 & 3.3 & 54.8 & 0.6 & 19.7 & 5.2 & \bf 8.3 \\
& RWKV (7B) & 2.9 & 28.1 & 4.4 & 26.6 & 4.4 & 26.6 & 5.1 & 25.9 & 69.0 & 4.1 & 22.5 & 6.6 & 33.4 & 0.1 & 10.4 & 6.5 & 29.6 \\
& Falcon (7B) & 6.0 & 61.4 & 8.5 & 58.9 & 8.4 & 59.0 & 9.5 & 57.9 & 32.6 & 18.7 & 59.1 & 7.6 & 72.5 & 5.4 & 84.5 & 8.3 & 44.1 \\
& Falcon (40B) & 7.7 & 44.0 & 9.7 & 42.1 & 9.7 & 42.1 & 10.8 & 41.0 & 48.2 & 23.9 & 39.7 & 5.8 & 53.5 & 8.1 & 65.3 & 8.8 & 29.5 \\
& Falcon-Instruct (7B) & 3.6 & 59.8 & 5.7 & 57.8 & 5.7 & 57.8 & 6.8 & 56.7 & 36.5 & 12.4 & 64.5 & 5.0 & 66.9 & 1.4 & 67.5 & 7.3 & 47.1 \\
& Falcon-Instruct (40B) & 7.8 & 35.2 & 10.2 & 32.8 & 10.1 & 32.9 & 10.8 & 32.2 & 57.0 & 26.0 & 34.5 & 6.0 & 50.9 & 8.0 & 63.1 & 8.2 & 15.3 \\

\bottomrule
\end{tabular}
\captionof{table}{Comparison of LLMs’ factuality about head, torso, and tail entities using all instances and instances of each domain from {\dn}. All numbers are in percentage (\%).}

\label{tab:mainresults}
\end{minipage}

\end{document}